\documentclass[compsoc, a4paper, 10 pt, conference]{IEEEtran}

\usepackage{etoolbox}
\makeatletter
\patchcmd{\@makecaption}
  {\scshape}
  {}
  {}
  {}
\makeatother     

\usepackage[font=small,labelfont=bf]{caption}
\usepackage{tabularx}
\usepackage{graphicx}
\usepackage{subcaption}
\usepackage{url}
\usepackage{calc}
\usepackage{amssymb}
\usepackage{amstext}
\usepackage{amsmath}
\usepackage[authoryear, nonamebreak, round, sort]{natbib}

\usepackage{xcolor}
\definecolor{dimgray}{rgb}{0.4 0.4 0.45}

\usepackage[]{hyperref}
\hypersetup{
    colorlinks=true,
    allcolors=dimgray,
    linkcolor=dimgray,
    filecolor=dimgray,      
    urlcolor=dimgray,
}

\usepackage[noabbrev, capitalise, nameinlink]{cleveref}

\usepackage{extdash}

\title{\LARGE \bf
Improving Unsupervised Defect Segmentation \\
by Applying Structural Similarity To Autoencoders
}

\author{\IEEEauthorblockN{Paul Bergmann$^{1}$,
Sindy L\"owe$^{1,2}$, Michael Fauser$^{1}$, David Sattlegger$^{1}$, and
Carsten Steger$^{1}$}
\vspace{0.5cm}
\IEEEauthorblockA{
$^{1}$MVTec Software GmbH\\
www.mvtec.com \\
\vspace{0.5cm}
\{bergmannp,fauser,sattlegger,steger\}@mvtec.com}
$^{2}$University of Amsterdam\\
sindy.lowe@student.uva.nl
}

\begin{document}

\maketitle
\thispagestyle{empty}
\pagestyle{empty}

%%%%%%%%%%%%%%%%%%%%%%%%%%%%%%%%%%%%%%%%%%%%%%%%%%%%%%%%%%%%%%%%%%%%%%%%%%%%%%%%
\begin{abstract}

Convolutional autoencoders have emerged as popular methods for unsupervised defect segmentation on image data. Most commonly, this task is performed by thresholding a per-pixel reconstruction error based on an $\ell^p$-distance. This procedure, however, leads to large residuals whenever the reconstruction includes slight localization inaccuracies around edges. It also fails to reveal defective regions that have been visually altered when intensity values stay roughly consistent. We show that these problems prevent these approaches from being applied to complex real-world scenarios and that they cannot be easily avoided by employing more elaborate architectures such as variational or feature matching autoencoders. We propose to use a perceptual loss function based on structural similarity that examines inter-dependencies between local image regions, taking into account luminance, contrast, and structural information, instead of simply comparing single pixel values. It achieves significant performance gains on a challenging real-world dataset of nanofibrous materials and a novel dataset of two woven fabrics over state-of-the-art approaches for unsupervised defect segmentation that use per-pixel reconstruction error metrics.

\end{abstract}

%%%%%%%%%%%%%%%%%%%%%%%%%%%%%%%%%%%%%%%%%%%%%%%%%%%%%%%%%%%%%%%%%%%%%%%%%%%%%%%%
\section{INTRODUCTION}

\noindent Visual inspection is essential in industrial manufacturing to ensure high production quality and high cost efficiency by quickly discarding defective parts. Since manual inspection by humans is slow, expensive, and error-prone, the use of fully automated computer vision systems is becoming increasingly popular. Supervised methods, where the system learns how to segment defective regions by training on both defective and non-defective samples, are commonly used. However, they involve a large effort to annotate data and all possible defect types need to be known beforehand. Furthermore, in some production processes, the scrap rate might be too small to produce a sufficient number of defective samples for training, especially for data-hungry deep learning models. 

% EYE CATCHER FIGURE
\begin{figure*}[t!]
\vspace{-0.3cm}
  \centering
    \centering
	\includegraphics[width=0.98\textwidth]{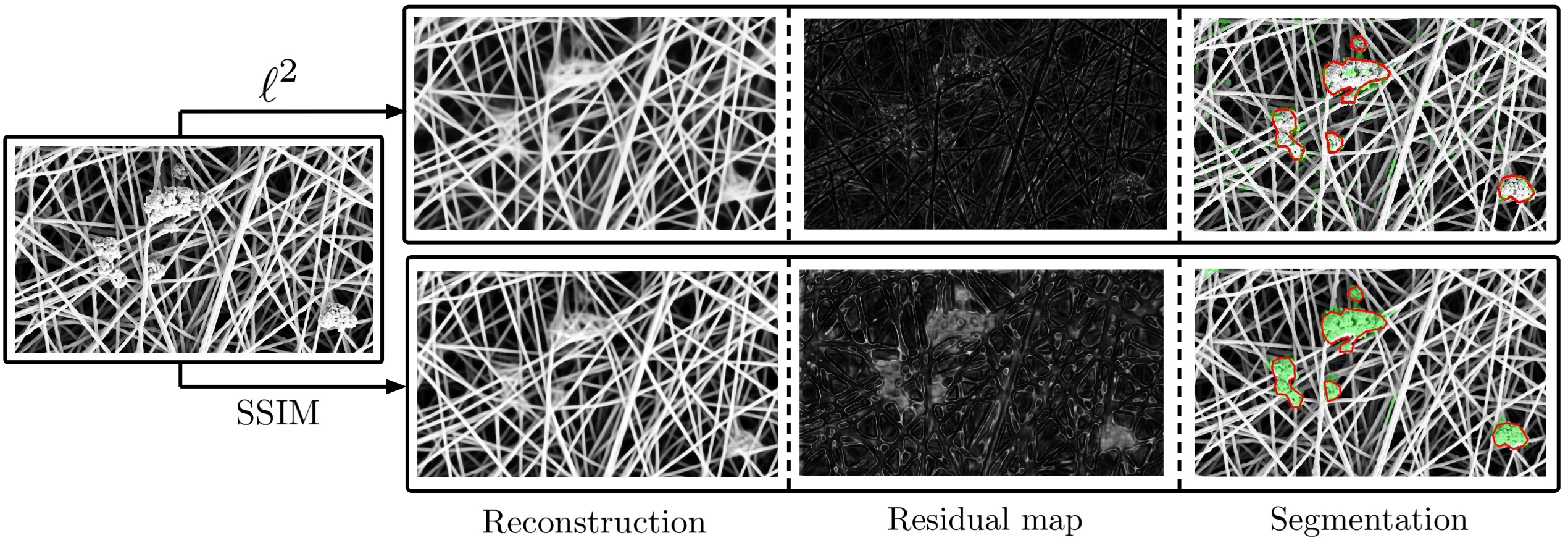}
  \caption{A defective image of nanofibrous materials is reconstructed by an autoencoder optimizing either the commonly used pixel-wise $\ell^2$-distance or a perceptual similarity metric based on structural similiarity (SSIM). Even though an $\ell^{2}$-autoencoder fails to properly reconstruct the defects, a per-pixel comparison of the original input and reconstruction does not yield significant residuals that would allow for defect segmentation. The residual map using SSIM puts more importance on the visually salient changes made by the autoencoder, enabling for an accurate segmentation of the defects.}
  \label{fig:eyecatcher}
\end{figure*}

In this work, we focus on unsupervised defect segmentation for visual inspection. The goal is to segment defective regions in images after having trained exclusively on non-defective samples. It has been shown that architectures based on convolutional neural networks (CNNs) such as autoencoders \citep{Goodfellow-et-al-2016} or generative adversarial networks (GANs; \citealp{goodfellow2014generative}) can be used for this task. We provide a brief overview of such methods in \cref{sec:related_work}. These models try to reconstruct their inputs in the presence of certain constraints such as a bottleneck and thereby manage to capture the essence of high-dimensional data (e.g., images) in a lower-dimensional space. It is assumed that anomalies in the test data deviate from the training data manifold and the model is unable to reproduce them. As a result, large reconstruction errors indicate defects. Typically, the error measure that is employed is a per-pixel $\ell^p$-distance, which is an ad-hoc choice made for the sake of simplicity and speed. However, these measures yield high residuals in locations where the reconstruction is only slightly inaccurate, e.g., due to small localization imprecisions of edges. They also fail to detect structural differences between the input and reconstructed images when the respective pixels' color values are roughly consistent. We show that this limits the usefulness of such methods when employed in complex real-world scenarios. 

To alleviate the aforementioned problems, we propose to measure reconstruction accuracy using the structural similarity (SSIM) metric \citep{ssim_index}. SSIM is a distance measure designed to capture perceptual similarity that is less sensitive to edge alignment and gives importance to salient differences between input and reconstruction. It captures inter-dependencies between local pixel regions that are disregarded by the current state-of-the-art unsupervised defect segmentation methods based on autoencoders with per-pixel losses. We evaluate the performance gains obtained by employing SSIM as a loss function on two real-world industrial inspection datasets and demonstrate significant performance gains over per-pixel approaches. \cref{fig:eyecatcher} demonstrates the advantage of perceptual loss functions over a per-pixel $\ell^2$-loss on the NanoTWICE dataset of nanofibrous materials \citep{handcrafted_feature_dictionary_nanofibres}. While both autoencoders alter the reconstruction in defective regions, only the residual map of the SSIM autoencoder allows a segmentation of these areas. By changing the loss function and otherwise keeping the same autoencoding architecture, we reach a performance that is on par with other state-of-the-art unsupervised defect segmentation approaches that rely on additional model priors such as handcrafted features or pretrained networks. 

\section{\texorpdfstring{\uppercase{Related Work}}{Related Work}}
\label{sec:related_work}

Detecting anomalies that deviate from the training data has been a long-standing problem in machine learning. \citet{pimentel2014review} give a comprehensive overview of the field. In computer vision, one needs to distinguish between two variants of this task. First, there is the classification scenario, where novel samples appear as entirely different object classes that should be predicted as outliers. Second, there is a scenario where anomalies manifest themselves in subtle deviations from otherwise known structures and a segmentation of these deviations is desired. For the classification problem, a number of approaches have been proposed \citep{perera_deep_features_one_class,sabokrou2018adversarially_OCC}. Here, we limit ourselves to an overview of methods that attempt to tackle the latter problem.
    
\citet{cnn_feature_dictionary_nanofibres} extract features from a CNN that has been pretrained on a classification task. The features are clustered in a dictionary during training and anomalous structures are identified when the extracted features strongly deviate from the learned cluster centers. General applicability of this approach is not guaranteed since the pretrained network might not extract useful features for the new task at hand and it is unclear which features of the network should be selected for clustering. The results achieved with this method are the current state-of-the-art on the NanoTWICE dataset, which we also use in our experiments. They improve upon previous results by \citet{handcrafted_feature_dictionary_nanofibres}, who build a dictionary that yields a sparse representation of the normal data. Similar approaches using sparse representations for novelty detection are \citep{sparse_representations_nanofibres,carrera_ano_conv_sparse,carrera2016scale}.

\citet{schlegl_anogan} train a GAN on optical coherence tomography images of the retina and detect anomalies such as retinal fluid by searching for a latent sample that minimizes the per-pixel $\ell^2$-reconstruction error as well as a discriminator loss. The large number of optimization steps that must be performed to find a suitable latent sample makes this approach very slow. Therefore, it is only useful in applications that are not time-critical. Recently, \citet{zenati2018BiGAN} proposed to use bidirectional GANs \citep{bigan} to add the missing encoder network for faster inference. However, GANs are prone to run into mode collapse, i.e., there is no guarantee that all modes of the distribution of non-defective images are captured by the model. Furthermore, they are more difficult to train than autoencoders since the loss function of the adversarial training typically cannot be trained to convergence \citep{arjovsky2017towards_training_gans}. Instead, the training results must be judged manually after regular optimization intervals. 

\citet{c_baur_vae_gan} propose a framework for defect segmentation using autoencoding architectures and a per-pixel error metric based on the $\ell^1$-distance. To prevent the disadvantages of their loss function, they improve the reconstruction quality by requiring aligned input data and adding an adversarial loss to enhance the visual quality of the reconstructed images. However, for many applications that work on unstructured data, prior alignment is impossible. Furthermore, optimizing for an additional adversarial loss during training but simply segmenting defects based on per-pixel comparisons during evaluation might lead to worse results since it is unclear how the adversarial training influences the reconstruction.

Other approaches take into account the structure of the latent space of variational autoencoders (VAEs; \citealp{kingma2013vae}) in order to define measures for outlier detection. \citet{vae_novelty_recon_probability} define a reconstruction probability for every image pixel by drawing multiple samples from the estimated encoding distribution and measuring the variability of the decoded outputs. \citet{soukupreliably_disregard_decoder} disregard the decoder output entirely and instead compute the KL divergence as a novelty measure between the prior and the encoder distribution. This is based on the assumption that defective inputs will manifest themselves in mean and variance values that are very different from those of the prior. Similarly, \citet{q_space_novelty} define multiple novelty measures, either by purely considering latent space behavior or by combining these measures with per-pixel reconstruction losses. They obtain a single scalar value that indicates an anomaly, which can quickly become a performance bottleneck in a segmentation scenario where a separate forward pass would be required for each image pixel to obtain an accurate segmentation result. We show that per-pixel reconstruction probabilities obtained from VAEs suffer from the same problems as per-pixel deterministic losses (cf.\ \cref{sec:experiments}). 

All the aforementioned works that use autoencoders for unsupervised defect segmentation have shown that autoencoders reliably reconstruct non-defective images while visually altering defective regions to keep the reconstruction close to the learned manifold of the training data. However, they rely on per-pixel loss functions that make the unrealistic assumption that neighboring pixel values are mutually independent. We show that this prevents these approaches from segmenting anomalies that differ predominantly in structure rather than pixel intensity. Instead, we propose to use SSIM \citep{ssim_index} as the loss function and measure of anomaly by comparing input and reconstruction. SSIM takes interdependencies of local patch regions into account and evaluates their first and second order moments to model differences in luminance, contrast, and structure. \citet{ae_vae_with_ms_ssim} show that SSIM and the closely related multi-scale version MS\Hyphdash*SSIM \citep{ms_ssim_index} can be used as differentiable loss functions to generate more realistic images in deep architectures for tasks such as superresolution, but do not examine its usefulness for defect segmentation in an autoencoding framework. In all our experiments, switching from per-pixel to perceptual losses yields significant gains in performance, sometimes enhancing the method from a complete failure to a satisfactory defect segmentation result.

\section{\texorpdfstring{\uppercase{Methodology}}{Methodology}}
\label{sec:methodology}

\subsection{Autoencoders for Unsupervised Defect Segmentation} 
\label{sec:autoencoders}

Autoencoders attempt to reconstruct an input image $\textbf{x} \in \mathbb{R}^{k \times h \times w}$ through a bottleneck, effectively projecting the input image into a lower-dimensional space, called latent space. An autoencoder consists of an encoder function $E: \mathbb{R}^{k \times h \times w} \rightarrow \mathbb{R}^{d}$ and a decoder function $D: \mathbb{R}^{d} \rightarrow \mathbb{R}^{k \times h \times w}$, where $d$ denotes the dimensionality of the latent space and $k,h,w$ denote the number of channels, height, and width of the input image, respectively. Choosing $d \ll k \times h \times w$ prevents the architecture from simply copying its input and forces the encoder to extract meaningful features from the input patches that facilitate accurate reconstruction by the decoder.  The overall process can be summarized as
\begin{equation}
  \hat{\textbf{x}} = D(E(\textbf{x})) = D(\textbf{z}) \enspace ,
  \label{eq:AE_equation}
\end{equation}
where $\textbf{z}$ is the latent vector and $\hat{\textbf{x}}$ the reconstruction of the input. In our experiments, the functions $E$ and $D$ are parameterized by CNNs. Strided convolutions are used to down-sample the input feature maps in the encoder and to up-sample them in the decoder. Autoencoders can be employed for unsupervised defect segmentation by training them purely on defect-free image data. During testing, the autoencoder will fail to reconstruct defects that have not been observed during training, which can thus be segmented by comparing the original input to the reconstruction and computing a residual map $R(\textbf{x},\hat{\textbf{x}}) \in \mathbb{R}^{w \times h}$.

\subsubsection{\texorpdfstring{$\ell^2$-Autoencoder}{l2-Autoencoder}}

To force the autoencoder to reconstruct its input, a loss function must be defined that guides it towards this behavior. For simplicity and computational speed, one often chooses a per-pixel error measure, such as the $L_2$ loss
\begin{equation}
  L_2(\textbf{x},\hat{\textbf{x}}) = \sum_{r=0}^{h-1} \sum_{c=0}^{w-1} {\left( \textbf{x}(r,c) - \hat{\textbf{x}}(r,c) \right)}^2 \enspace ,
\end{equation}
where $\textbf{x}(r,c)$ denotes the intensity value of image $\textbf{x}$ at the pixel $(r,c)$. To obtain a residual map $R_{\ell^2}(\textbf{x},\hat{\textbf{x}})$ during evaluation, the per-pixel $\ell^2$-distance of $\textbf{x}$ and $\hat{\textbf{x}}$ is computed.

 \begin{figure}[t!]
  \centering
    %\vspace{1.3cm}
    \centering
    \includegraphics[width=0.45\textwidth]{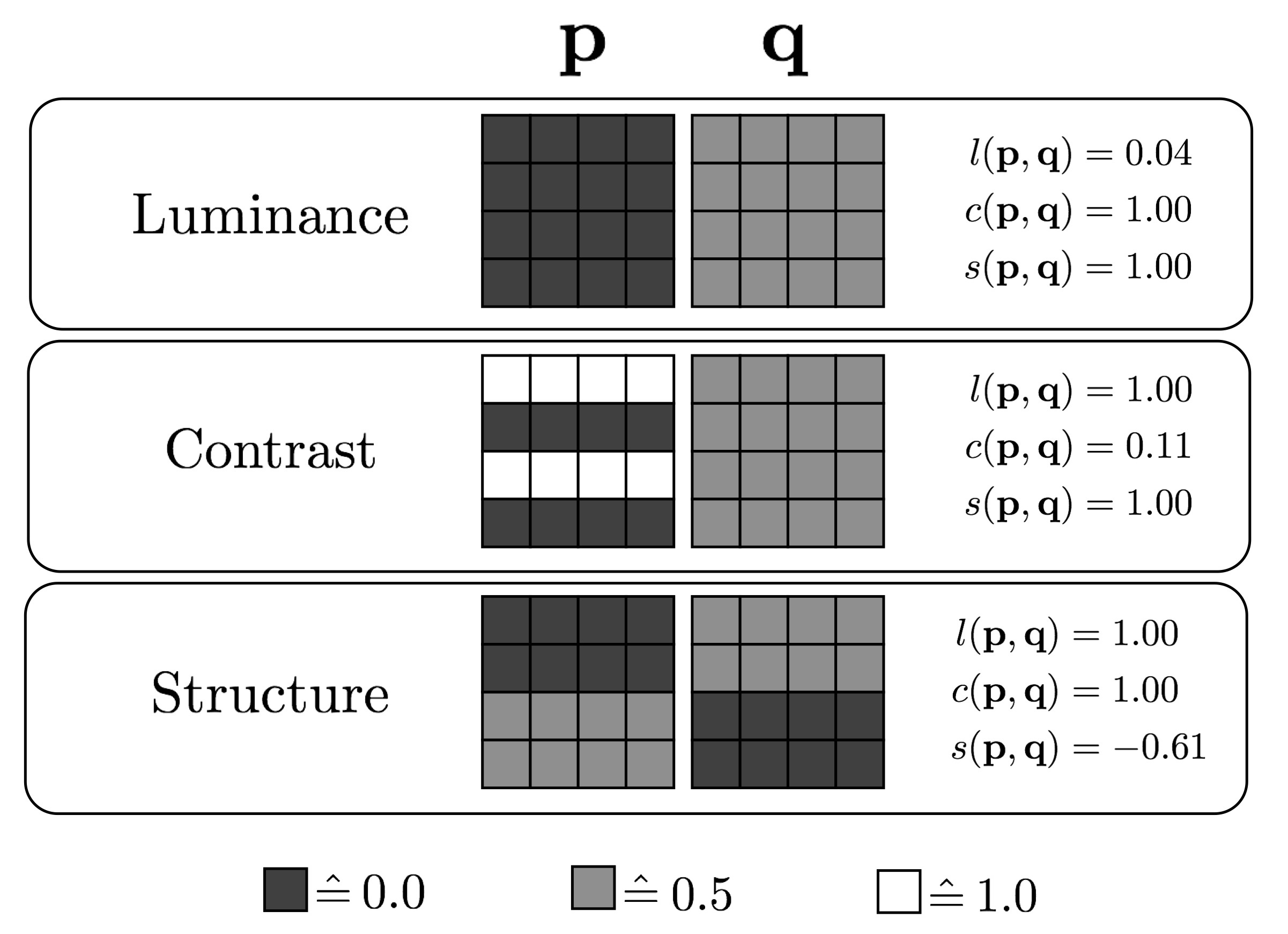}
  \caption{Different responsibilities of the three similarity functions employed by SSIM. Example patches $\textbf{p}$ and $\textbf{q}$ differ in either luminance, contrast, or structure. SSIM is able to distinguish between these three cases, assigning close to minimum similarity values to one of the comparison functions $l(\textbf{p},\textbf{q})$, $c(\textbf{p},\textbf{q})$, or $s(\textbf{p},\textbf{q})$, respectively. An $\ell^2$-comparison of these patches would yield a constant per-pixel residual value of 0.25 for each of the three cases.}
  \label{fig:ssim_patch_calculation}
\end{figure}

\begin{figure*}[t!]
  \vspace{-0.3cm}
  \centering    
    \centering
    \begin{subfigure}[c]{0.14\textwidth}
    \centering
	\includegraphics[width=0.99\textwidth]{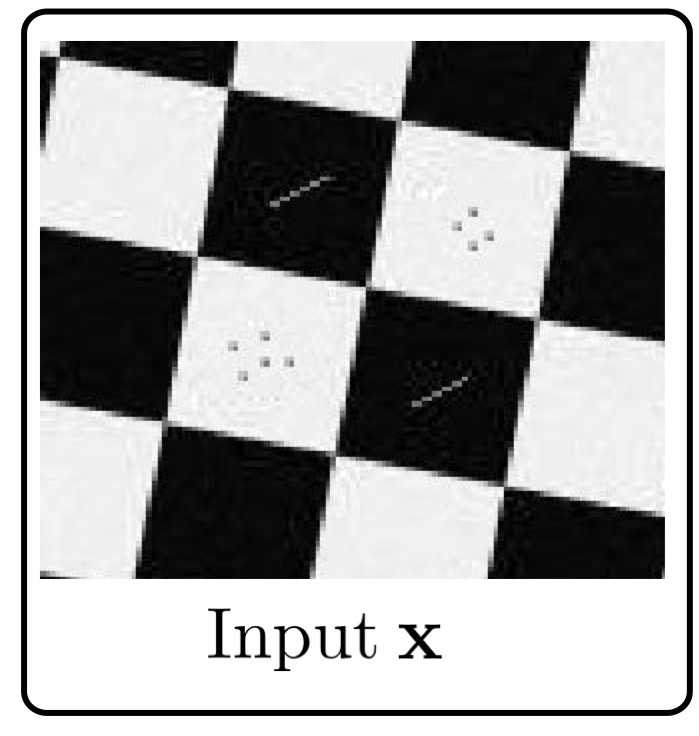}
	\subcaption{}
    \label{fig:checkerboard_a}
  \end{subfigure}
  \begin{subfigure}[c]{0.14\textwidth}
    \centering
	\includegraphics[width=0.99\textwidth]{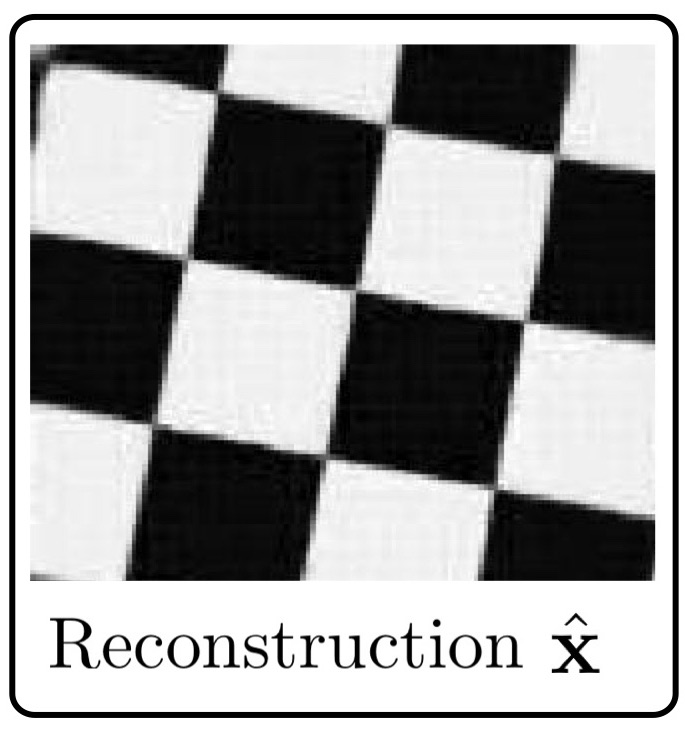}
	\subcaption{}
    \label{fig:checkerboard_b}
  \end{subfigure}
  \begin{subfigure}[c]{0.14\textwidth}
    \centering
	\includegraphics[width=2.23cm,height=2.4cm]{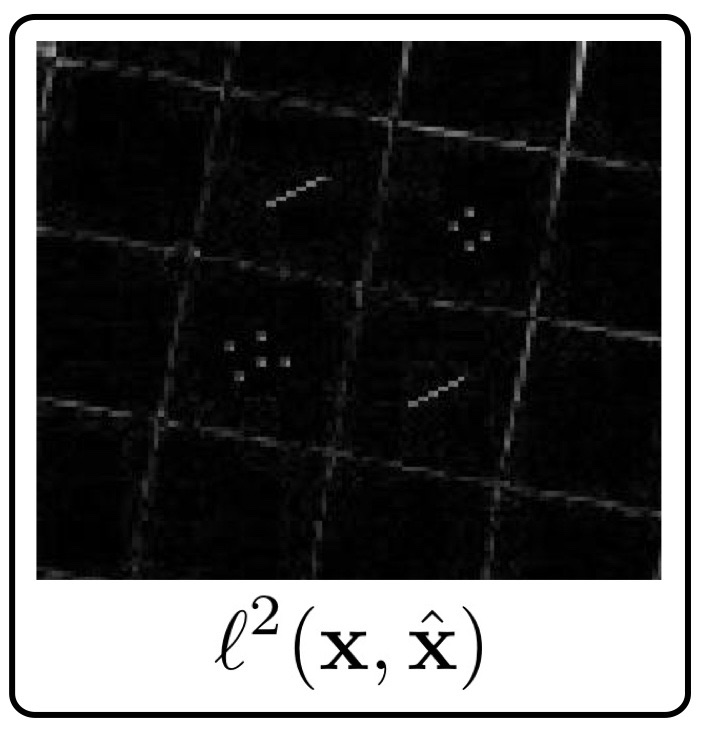}
	\subcaption{}
    \label{fig:checkerboard_c}
  \end{subfigure}
  \begin{subfigure}[c]{0.56\textwidth}
    \centering
	\includegraphics[width=0.99\textwidth]{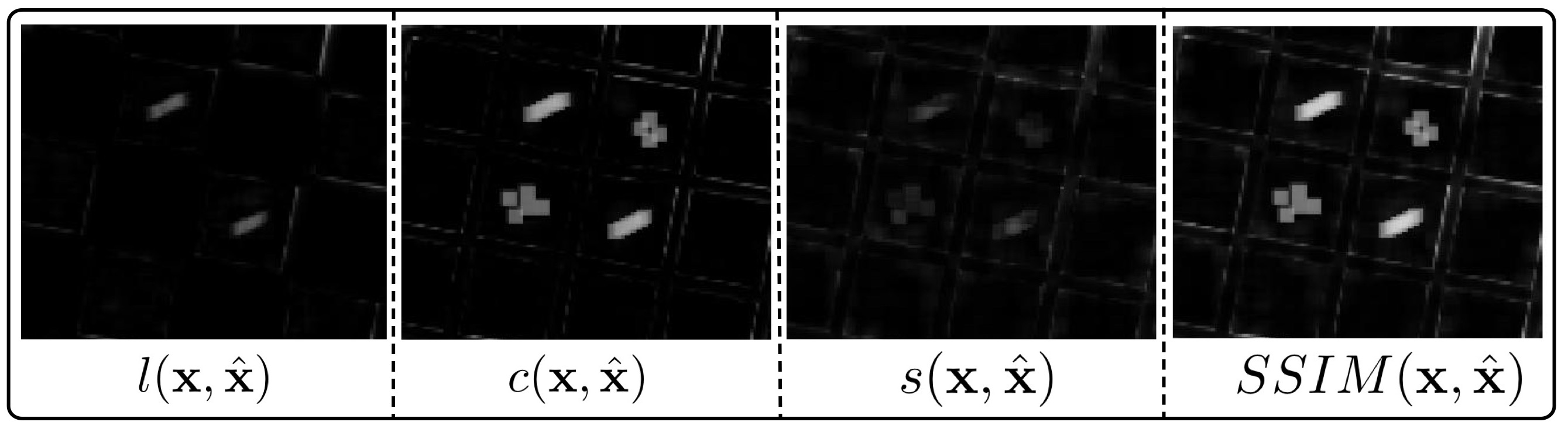}
	\subcaption{}
    \label{fig:checkerboard_d}
  \end{subfigure}
  \caption{A toy example illustrating the advantages of SSIM over $\ell^2$ for the segmentation of defects. \textbf{(a)} $128 \times 128$ checkerboard pattern with gray strokes and dots that simulate defects. \textbf{(b)} Output reconstruction $\hat{\textbf{x}}$ of the input image $\textbf{x}$ by an $\ell^2$-autoencoder trained on defect-free checkerboard patterns. The defects have been removed by the autoencoder. \textbf{(c)} $\ell^2$-residual map. Brighter colors indicate larger dissimilarity between input and reconstruction. \textbf{(d)} Residuals for luminance $l$, contrast $c$, structure $s$, and their pointwise product that yields the final SSIM residual map. In contrast to the $\ell^2$-error map, SSIM gives more importance to the visually more salient disturbances than to the slight inaccuracies around reconstructed edges.}
   \label{fig:checkerboard}
\end{figure*}

\subsubsection{Variational Autoencoder}

Various extensions to the deterministic autoencoder framework exist. VAEs \citep{kingma2013vae} impose constraints on the latent variables to follow a certain distribution $\textbf{z} \sim P(\textbf{z})$. For simplicity, the distribution is typically chosen to be a unit-variance Gaussian. This turns the entire framework into a probabilistic model that enables efficient posterior inference and allows to generate new data from the training manifold by sampling from the latent distribution. The approximate posterior distribution $Q(\textbf{z}|\textbf{x})$ obtained by encoding an input image can be used to define further anomaly measures. One option is to compute a distance between the two distributions, such as the KL-divergence $\mathcal{KL}(Q(\textbf{z}|\textbf{x})||P(\textbf{z}))$, and indicate defects for large deviations from the prior $P(\textbf{z})$ \citep{soukupreliably_disregard_decoder}. However, to use this approach for the pixel-accurate segmentation of anomalies, a separate forward pass for each pixel of the input image would have to be performed. A second approach for utilizing the posterior $Q(\textbf{z}|\textbf{x})$ that yields a spatial residual map is to decode $N$ latent samples $\textbf{z}_1,\textbf{z}_2,\dots,\textbf{z}_N$ drawn from $Q(\textbf{z}|\textbf{x})$ and to evaluate the per-pixel reconstruction probability $R_{VAE} = P(\textbf{x}|\textbf{z}_1,\textbf{z}_2,\dots,\textbf{z}_N)$ as described by \citet{vae_novelty_recon_probability}.

\subsubsection{Feature Matching Autoencoder}

Another extension to standard autoencoders was proposed by \citet{brox_learned_visual_similarity_metrics}. It increases the quality of the produced reconstructions by extracting features from both the input image $\textbf{x}$ and its reconstruction $\hat{\textbf{x}}$ and enforcing them to be equal. Consider $F: \mathbb{R}^{k \times h \times w} \rightarrow \mathbb{R}^{f}$ to be a feature extractor that obtains an $f$-dimensional feature vector from an input image. Then, a regularizer can be added to the loss function of the autoencoder, yielding the feature matching autoencoder (FM-AE) loss
\begin{equation}
  L_{\textrm{FM}}(\textbf{x},\hat{\textbf{x}}) = L_2(\textbf{x},\hat{\textbf{x}}) + \lambda \| F(\textbf{x}) - F(\hat{\textbf{x}})\|_{2}^{2} \enspace , 
\label{eq:fm}
\end{equation}
where $\lambda > 0$ denotes the weighting factor between the two loss terms. $F$ can be parameterized using the first layers of a CNN pretrained on an image classification task. During evaluation, a residual map $R_\mathrm{FM}$ is obtained by comparing the per-pixel $\ell^2$-distance of $\textbf{x}$ and $\hat{\textbf{x}}$. The hope is that sharper, more realistic reconstructions will lead to better residual maps compared to a standard $\ell^2$-autoencoder.

\subsubsection{SSIM Autoencoder}

We show that employing more elaborate architectures such as VAEs or FM-AEs does not yield satisfactory improvements of the residial maps over deterministic $\ell^2$-autoencoders in the unsupervised defect segmentation task. They are all based on per-pixel evaluation metrics that assume an unrealistic independence between neighboring pixels. Therefore, they fail to detect structural differences between the inputs and their reconstructions. By adapting the loss and evaluation functions to capture local inter-dependencies between image regions, we are able to drastically improve upon all the aforementioned architectures.  In \cref{sec:ssim}, we specifically motivate the use of the strucutural similarity metric SSIM$(\textbf{x},\hat{\textbf{x}})$ as both the loss function and the evaluation metric for autoencoders to obtain a residual map $R_{SSIM}$.

\subsection{Structural Similarity}
\label{sec:ssim}

The SSIM index \citep{ssim_index} defines a distance measure between two $K \times K$ image patches $\textbf{p}$ and $\textbf{q}$, taking into account their similarity in luminance $l(\textbf{p},\textbf{q})$, contrast $c(\textbf{p},\textbf{q})$, and structure $s(\textbf{p},\textbf{q})$:
\begin{equation}
  \mathrm{SSIM}(\textbf{p},\textbf{q}) = l(\textbf{p},\textbf{q})^\alpha c(\textbf{p},\textbf{q})^\beta s(\textbf{p},\textbf{q})^\gamma \enspace ,
  \label{eq:SSIM}
\end{equation}
where $\alpha, \beta, \gamma \in \mathbb{R}$ are user-defined constants to weight the three terms. The luminance measure $l(\textbf{p},\textbf{q})$ is estimated by comparing the patches' mean intensities $\mu_{\textbf{p}}$ and $\mu_{\textbf{q}}$. The contrast measure $c(\textbf{p},\textbf{q})$ is a function of the patch variances $\sigma_{\textbf{p}}^2$ and $\sigma_{\textbf{q}}^2$. The structure measure $s(\textbf{p},\textbf{q})$ takes into account the covariance $\sigma_{\textbf{p}\textbf{q}}$ of the two patches. The three measures are defined as:
\begin{eqnarray}
  \label{eq:l_comp}
  l(\textbf{p},\textbf{q}) & = & \frac{2 \mu_{\textbf{p}} \mu_{\textbf{q}} + c_1}{\mu_{\textbf{p}}^2 + \mu_{\textbf{q}}^2 + c_1} \\
  \label{eq:c_comp}
  c(\textbf{p},\textbf{q}) & = & \frac{2 \sigma_{\textbf{p}} \sigma_{\textbf{q}} + c_2}{\sigma_{\textbf{p}}^2 + \sigma_{\textbf{q}}^2 + c_2} \\
  \label{eq:s_comp}
  s(\textbf{p},\textbf{q}) & = & \frac{2 \sigma_{\textbf{p}\textbf{q}} + c_2}{2 \sigma_{\textbf{p}}\sigma_{\textbf{q}} + c_2} \enspace .
\end{eqnarray}
The constants $c_1$ and $c_2$ ensure numerical stability and are typically set to $c_1 = 0.01$ and $c_2 = 0.03$.  By substituting (\ref{eq:l_comp})-(\ref{eq:s_comp}) into (\ref{eq:SSIM}), the SSIM is given by
\begin{equation}
  \mathrm{SSIM}(\textbf{p},\textbf{q}) = \frac{(2 \mu_{\textbf{p}} \mu_{\textbf{q}} + c_1)(2 \sigma_{\textbf{p} \textbf{q}} + c_2)}{(\mu_{\textbf{p}}^2 + \mu_{\textbf{q}}^2+c_1)(\sigma_{\textbf{p}}^2 + \sigma_{\textbf{q}}^2 + c_2)} \enspace . 
  \label{ssimeq}
\end{equation}
It holds that $\mathrm{SSIM}(\textbf{p},\textbf{q}) \in [-1,1]$. In particular, $\mathrm{SSIM}(\textbf{p},\textbf{q}) = 1$ if and only if $\textbf{p}$ and $\textbf{q}$ are identical \citep{ssim_index}. \cref{fig:ssim_patch_calculation} shows the different perceptions of the three similarity functions that form the SSIM index. Each of the patch pairs $\textbf{p}$ and $\textbf{q}$ has a constant $\ell^2$-residual of 0.25 per pixel and hence assigns low defect scores to each of the three cases. SSIM on the other hand is sensitive to variations in the patches' mean, variance, and covariance in its respective residual map and assigns low similarity to each of the patch pairs in one of the comparison functions. 

To compute the structural similarity between an entire image $\textbf{x}$ and its reconstruction $\hat{\textbf{x}}$, one slides a $K \times K$ window across the image and computes a SSIM value at each pixel location. Since (\ref{ssimeq}) is differentiable, it can be employed as a loss function in deep learning architectures that are optimized using gradient descent. 

\Cref{fig:checkerboard} indicates the advantages SSIM has over per-pixel error functions such as $\ell^2$ for segmenting defects. After training an $\ell^2$-autoencoder on defect-free checkerboard patterns of various scales and orientations, we apply it to an image (\cref{fig:checkerboard_a}) that contains gray strokes and dots that simulate defects. \cref{fig:checkerboard_b} shows the corresponding reconstruction produced by the autoencoder, which removes the defects from the input image. The two remaining subfigures display the residual maps when evaluating the reconstruction error with a per-pixel $\ell^2$-comparison or SSIM. For the latter, the luminance, contrast, and structure maps are also shown. For the $\ell^2$-distance, both the defects and the inaccuracies in the reconstruction of the edges are weighted equally in the error map, which makes them indistinguishable. Since SSIM computes three different statistical features for image comparison and operates on local patch regions, it is less sensitive to small localization inaccuracies in the reconstruction. In addition, it detects defects that manifest themselves in a change of structure rather than large differences in pixel intensity. For the defects added in this particular toy example, the contrast function yields the largest residuals.

\section{\texorpdfstring{\uppercase{Experiments}}{Experiments}}
\label{sec:experiments}

\subsection{Datasets}
\label{sec:dataset}

\begin{figure}[t!]
 \vspace{-0.5cm}
  \centering    
  \begin{subfigure}[c]{0.23\textwidth}
    \centering
	\includegraphics[width=0.9\textwidth]{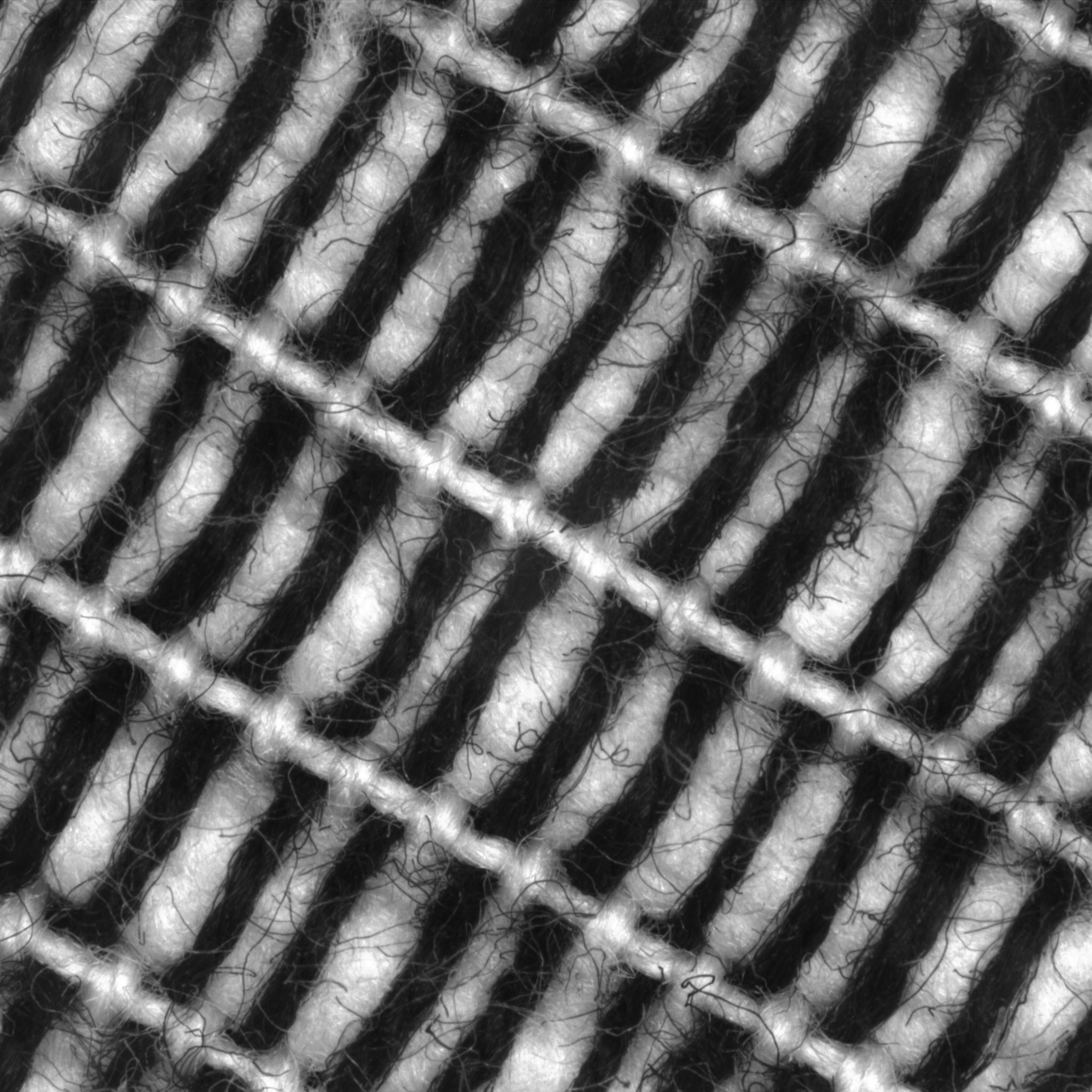}
	\subcaption{}
    \label{fig:texture_2_good}
  \end{subfigure}
  \begin{subfigure}[c]{0.23\textwidth}
    \centering
	\includegraphics[width=0.9\textwidth]{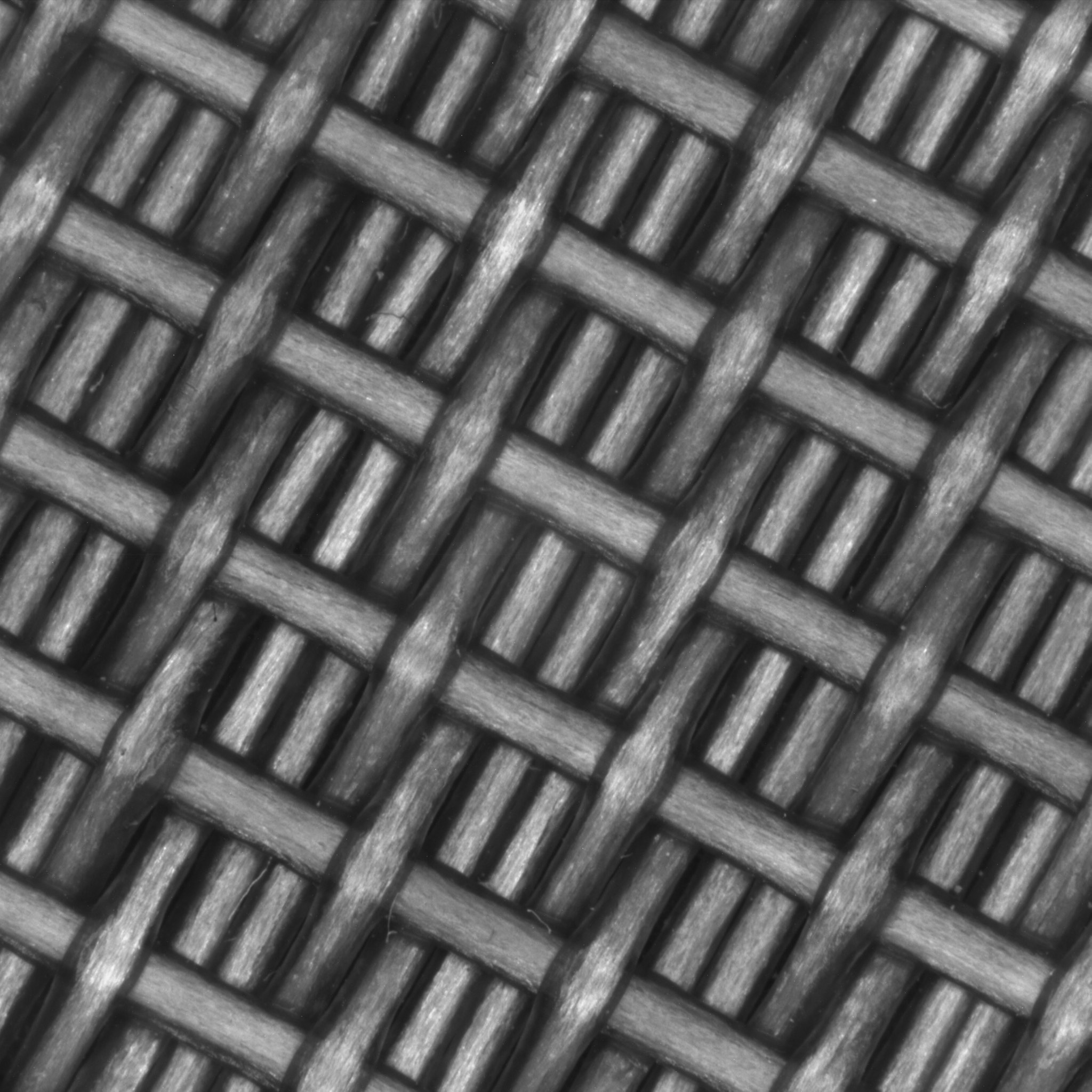}
	\subcaption{}
    \label{fig:texture_1_good}
  \end{subfigure}
  \begin{subfigure}[c]{0.23\textwidth}
    \centering
    \vspace{0.1cm}
	\includegraphics[width=0.9\textwidth]{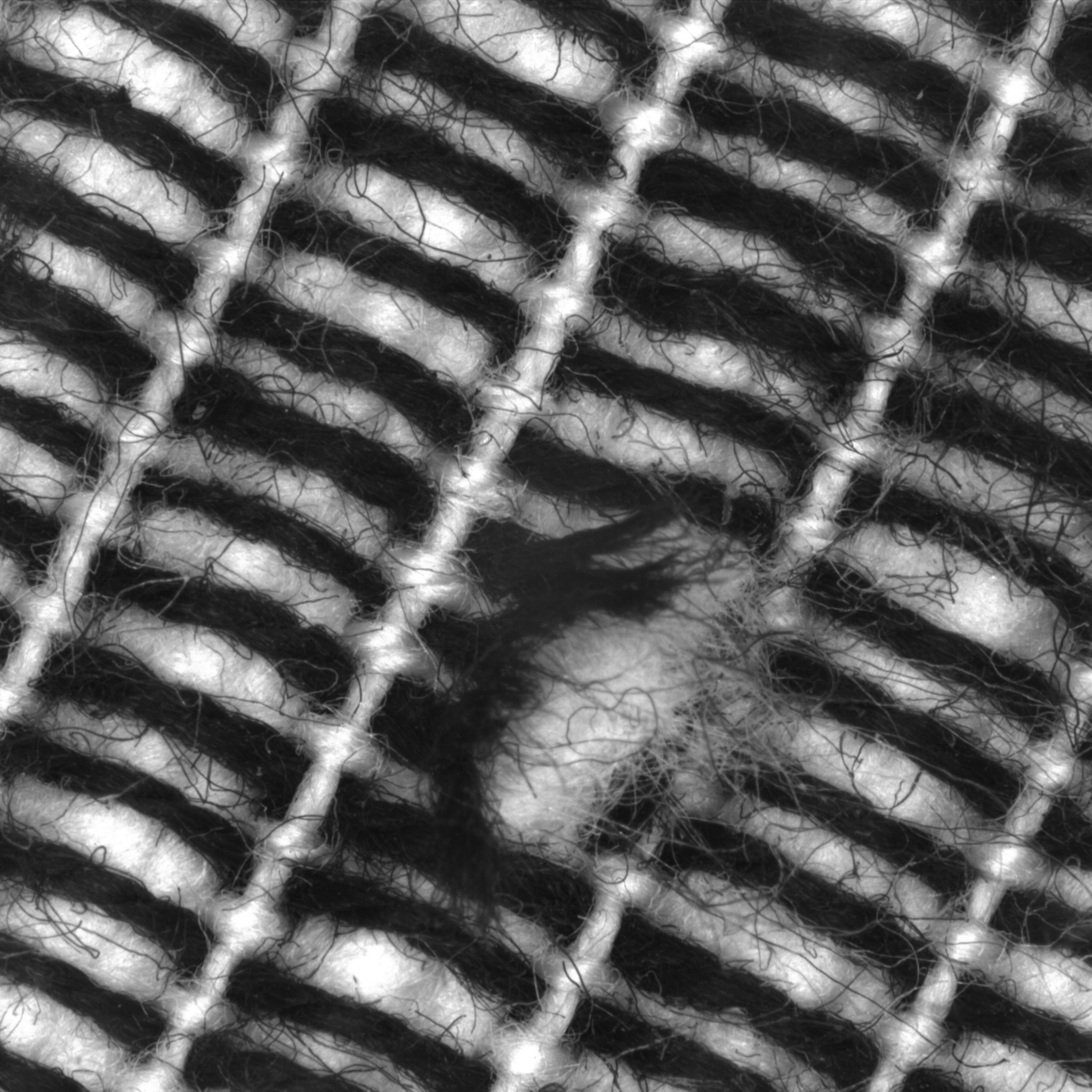}
	\subcaption{}
    \label{fig:texture_2_bad}
  \end{subfigure} 
  \begin{subfigure}[c]{0.23\textwidth}
    \centering
    \vspace{0.1cm}
	\includegraphics[width=0.9\textwidth]{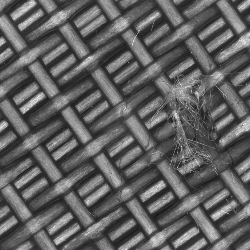}
	\subcaption{}
    \label{fig:texture_1_bad}
  \end{subfigure} 
  \caption{Example images from the contributed texture dataset of two woven fabrics. \textbf{(a)} and \textbf{(b)} show examples of non-defective textures that can be used for training. \textbf{(c)} and \textbf{(d)} show exemplary defects for both datasets. See the text for details.}

   \label{fig:texture_dataset}
\end{figure}

Due to the lack of datasets for unsupervised defect segmentation in industrial scenarios, we contribute a novel dataset of two woven fabric textures, which is available to the public\footnote{The dataset is available at \texttt{\url{https://www.mvtec.com/company/research/publications}}}. We provide 100 defect-free images per texture for training and validation and 50 images that contain various defects such as cuts, roughened areas, and contaminations on the fabric. Pixel-accurate ground truth annotations for all defects are also provided. All images are of size $\textrm{512} \times \textrm{512}$ pixels and were acquired as single-channel gray-scale images. Examples of defective and defect-free textures can be seen in \cref{fig:texture_dataset}. We further evaluate our method on a dataset of nanofibrous materials \citep{handcrafted_feature_dictionary_nanofibres}, which contains five defect-free gray-scale images of size $\textrm{1024} \times \textrm{700}$ for training and validation and 40 defective images for evaluation. A sample image of this dataset is shown in \cref{fig:eyecatcher}. 

\begin{table}[t!]
\centering
\scriptsize
\def\arraystretch{0.9}
\begin{tabular}{lcccc}
\hline
Layer & \multicolumn{1}{l}{Output Size} & \multicolumn{3}{c}{Parameters}                            \\
               & \multicolumn{1}{l}{}                     & Kernel      & Stride      & Padding     \\ \hline
Input          & 128x128x1                                & \multicolumn{1}{l}{} & \multicolumn{1}{l}{} & \multicolumn{1}{l}{} \\
Conv1          & 64x64x32                                 & 4x4                  & 2                    & 1                    \\
Conv2          & 32x32x32                                 & 4x4                  & 2                    & 1                    \\
Conv3          & 32x32x32                                 & 3x3                  & 1                    & 1                    \\
Conv4          & 16x16x64                                 & 4x4                  & 2                    & 1                    \\
Conv5          & 16x16x64                                 & 3x3                  & 1                    & 1                    \\
Conv6          & 8x8x128                                  & 4x4                  & 2                    & 1                    \\
Conv7          & 8x8x64                                   & 3x3                  & 1                    & 1                    \\
Conv8          & 8x8x32                                   & 3x3                  & 1                    & 1                    \\
Conv9          & 1x1x$d$                                  & 8x8                  & 1                    & 0                    \\ \hline
\end{tabular}
\caption{General outline of our autoencoder architecture. The depicted values correspond to the structure of the encoder. The decoder is built as a reversed version of this. Leaky rectified linear units (ReLUs) with slope 0.2 are applied as activation functions after each layer except for the output layers of both the encoder and the decoder, in which linear activation functions are used.}
\label{table:ae_arch}
\end{table}

\begin{figure*}[t!]
  \centering
  \includegraphics[width=0.95\textwidth]{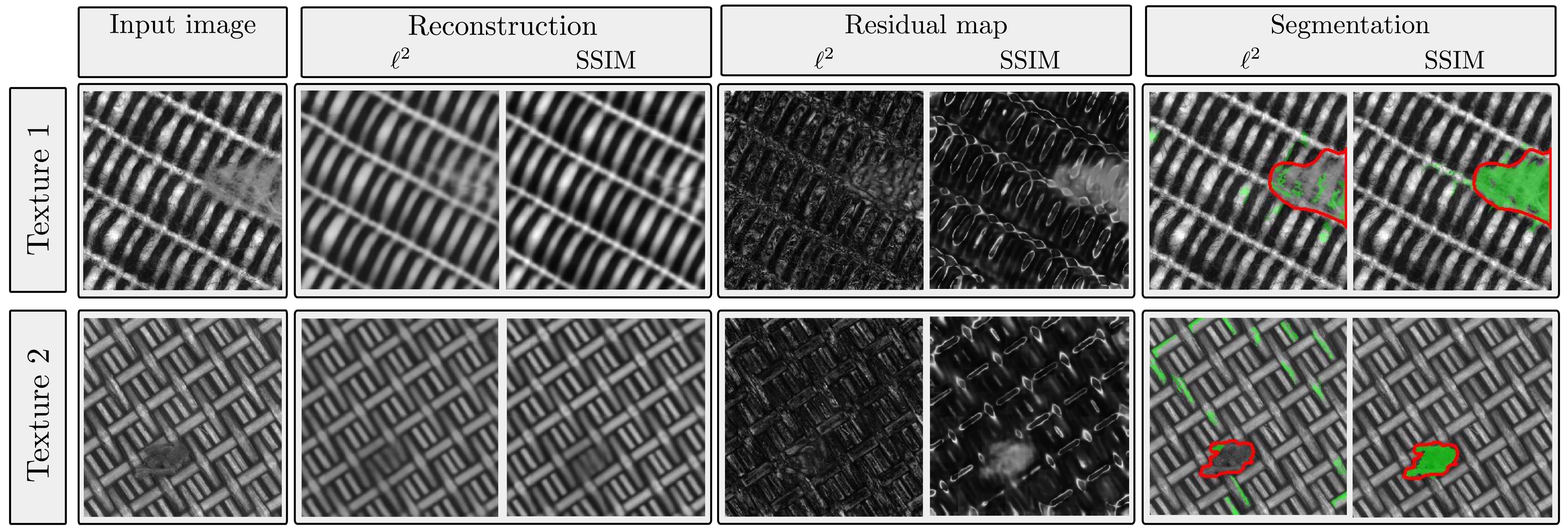}
  \caption{Qualitative comparison between reconstructions, residual maps, and segmentation results of an $\ell^2$-autoencoder and an SSIM autoencoder on two datasets of woven fabric textures. The ground truth regions containing defects are outlined in red while green areas mark the segmentation result of the respective method.}
  \label{fig:textures_qual_results}
\end{figure*}

\begin{figure*}[t!]
  \centering    
  \begin{subfigure}[c]{0.32\textwidth}
    \centering
	\includegraphics[width=0.98\textwidth]{Images/roc_archs_nano.pdf}
	\subcaption{}
    \label{fig:roc_main_results_a}
  \end{subfigure}
  \begin{subfigure}[c]{0.32\textwidth}
    \centering
	\includegraphics[width=0.98\textwidth]{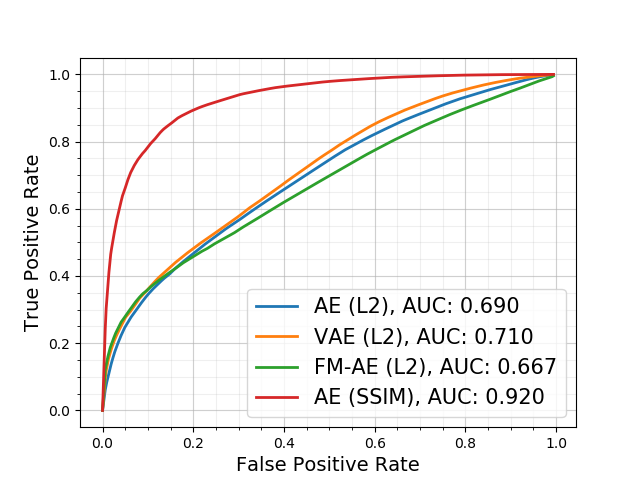}
	\subcaption{}
    \label{fig:roc_main_results_b}
  \end{subfigure}
  \begin{subfigure}[c]{0.32\textwidth}
    \centering
	\includegraphics[width=0.98\textwidth]{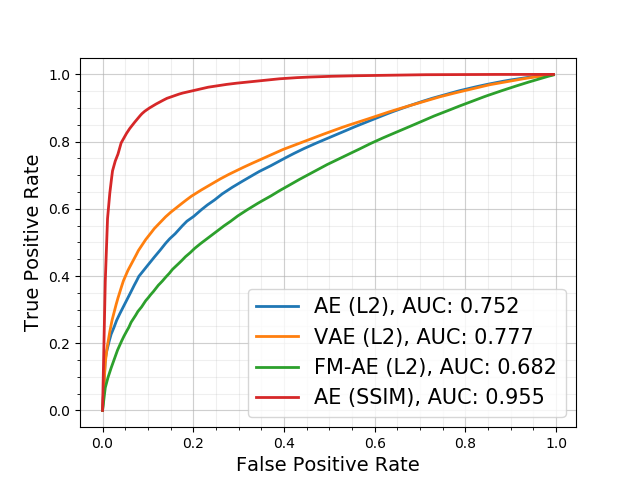}
	\subcaption{}
    \label{fig:roc_main_results_c}
  \end{subfigure}
  \caption{Resulting ROC curves of the proposed SSIM autoencoder (red line) on the evaluated datasets of nanofibrous materials \textbf{(a)} and the two texture datasets \textbf{(b}), (\textbf{c)} in comparison with other autoencoding architectures that use per-pixel loss functions (green, orange, and blue lines). Corresponding AUC values are given in the legend.} 
   \label{fig:roc_main_results}
\end{figure*}

\subsection{Training and Evaluation Procedure}

For all datasets, we train the autoencoders with their respective losses and evaluation metrics, as described in \cref{sec:autoencoders}. Each architecture is trained on 10\,000 defect-free patches of size $\textrm{128} \times \textrm{128}$, randomly cropped from the given training images. In order to capture a more global context of the textures, we down-scaled the images to size $\textrm{256} \times \textrm{256}$ before cropping. Each network is trained for 200 epochs using the ADAM \citep{kingma2014adam} optimizer with an initial learning rate of $2 \times 10^{-4}$ and a weight decay set to $10^{-5}$. The exact parametrization of the autoencoder network shared by all tested architectures is given in \cref{table:ae_arch}. The latent space dimension for our experiments is set to $d = 100$ on the texture images and to $d = 500$ for the nanofibres due to their higher structural complexity. For the VAE, we decode $N=6$ latent samples from the approximate posterior distribution $Q(\textbf{z}|\textbf{x})$ to evaluate the reconstruction probability for each pixel. The feature matching autoencoder is regularized with the first three convolutional layers of an AlexNet \citep{krizhevsky2012imagenet} pretrained on ImageNet \citep{russakovsky2015imagenet} and a weight factor of $\lambda = 1$. For SSIM, the window size is set to $K = 11$ unless mentioned otherwise and its three residual maps are equally weighted by setting $\alpha = \beta = \gamma = 1$.

The evaluation is performed by striding over the test images and reconstructing image patches of size $\textrm{128} \times \textrm{128}$ using the trained autoencoder and computing its respective residual map $R$. In principle, it would be possible to set the horizontal and vertical stride to 128. However, at different spatial locations, the autoencoder produces slightly different reconstructions of the same data, which leads to some striding artifacts. Therefore, we decreased the stride to 30 pixels and averaged the reconstructed pixel values. The resulting residual maps are thresholded to obtain candidate regions where a defect might be present. An opening with a circular structuring element of diameter~4 is applied as a morphological post-processing to delete outlier regions that are only a few pixels wide \citep{machine_vision_algorithms_and_applications}. We compute the receiver operating characteristic (ROC) as the evaluation metric. The true positive rate is defined as the ratio of pixels correctly classified as defect across the entire dataset. The false positive rate is the ratio of pixels misclassified as defect.

\subsection{Results}

%(1) Qualitative results on the texture dataset (and also teaser figure)
\cref{fig:textures_qual_results} shows a qualitative comparison between the performance of the $\ell^2$-autoencoder and the SSIM autoencoder on images of the two texture datasets. Although both architectures remove the defect in the reconstruction, only the SSIM residual map reveals the defects and provides an accurate segmentation result. The same can be observed for the NanoTWICE dataset, as shown in \cref{fig:eyecatcher}. 

%(2) Quantitative results in terms of ROC curves for all 3 datasets
We confirm this qualitative behavior by numerical results. \Cref{fig:roc_main_results} compares the ROC curves and their respective AUC values of our approach using SSIM to the per-pixel architectures. The performance of the latter is often only marginally better than classifying each pixel randomly. For the VAE, we found that the reconstructions obtained by different latent samples from the posterior does not vary greatly. Thus, it could not improve on the deterministic framework. Employing feature matching only improved the segmentation result for the dataset of nanofibrous materials, while not yielding a benefit for the two texture datasets. Using SSIM as the loss and evaluation metric outperforms all other tested architectures significantly. By merely changing the loss function, the achieved AUC improves from 0.688 to 0.966 on the dataset of nanofibrous materials, which is comparable to the state-of-the-art by \citet{cnn_feature_dictionary_nanofibres}, where values of up to 0.974 are reported. In contrast to this method, autoencoders do not rely on any model priors such as handcrafted features or pretrained networks. For the two texture datasets, similar leaps in performance are observed. 

\begin{figure}[t!]
    \vspace{-0.3cm}
  \centering    
    \centering
	\includegraphics[width=0.45\textwidth]{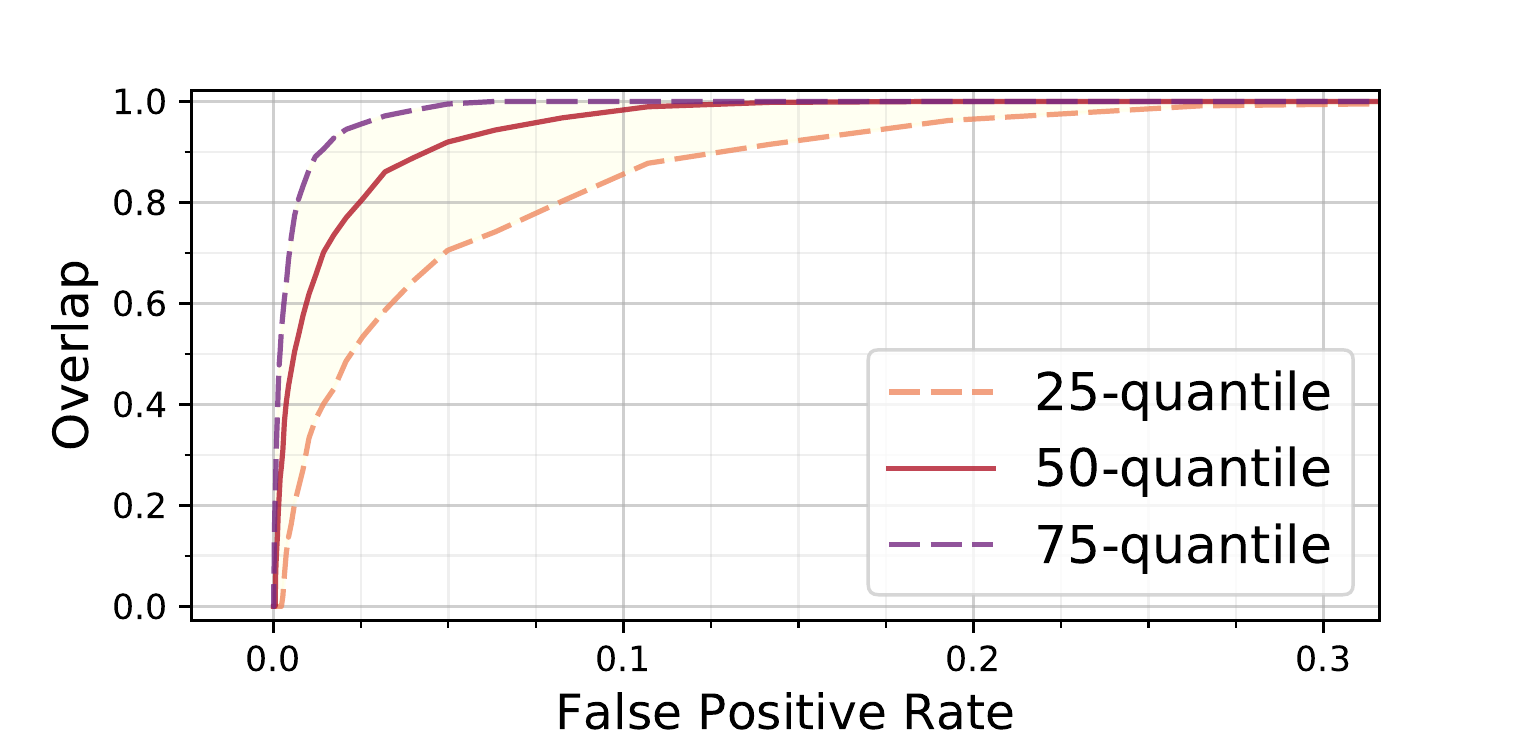}
  \caption{Per-region overlap for individual defects between our segmentation and the ground truth for different false positive rates using an SSIM autoencoder on the dataset of nanofibrous materials.}
   \label{fig:quantiles}
\end{figure}

Since the dataset of nanofibrous materials contains defects of various sizes and smaller sized defects contribute less to the overall true positive rate when weighting all pixel equally, we further evaluated the overlap of each detected anomaly region with the ground truth for this dataset and report the $p$-quantiles for $p \in \{25\%,50\%,75\%\}$ in \cref{fig:quantiles}. For false positive rates as low as 5\%, more than 50\% of the defects have an overlap with the ground truth that is larger than 91\%. This outperforms the results achieved by \citet{cnn_feature_dictionary_nanofibres}, who report a minimal overlap of 85\% in this setting. 

We further tested the sensitivity of the SSIM autoencoder to different hyperparameter settings. We varied the latent space dimension $d$, SSIM window size $k$, and the size of the patches that the autoencoder was trained on. \cref{table:hyperparams} shows that SSIM is insensitive to different hyperparameter settings once the latent space dimension is chosen to be sufficiently large. Using the optimal setup of $d=500$, $k=11$, and patch size $\textrm{128} \times \textrm{128}$, a forward pass through our architecture takes 2.23\,ms on a Tesla V100 GPU. Patch-by-patch evaluation of an entire image of the NanoTWICE dataset takes 3.61\,s on average, which is significantly faster than the runtimes reported by \citet{cnn_feature_dictionary_nanofibres}. Their approach requires between 15\,s and 55\,s to process a single input image.

\begin{figure}[t!]
  \centering    
    \centering
	\includegraphics[width=0.45\textwidth]{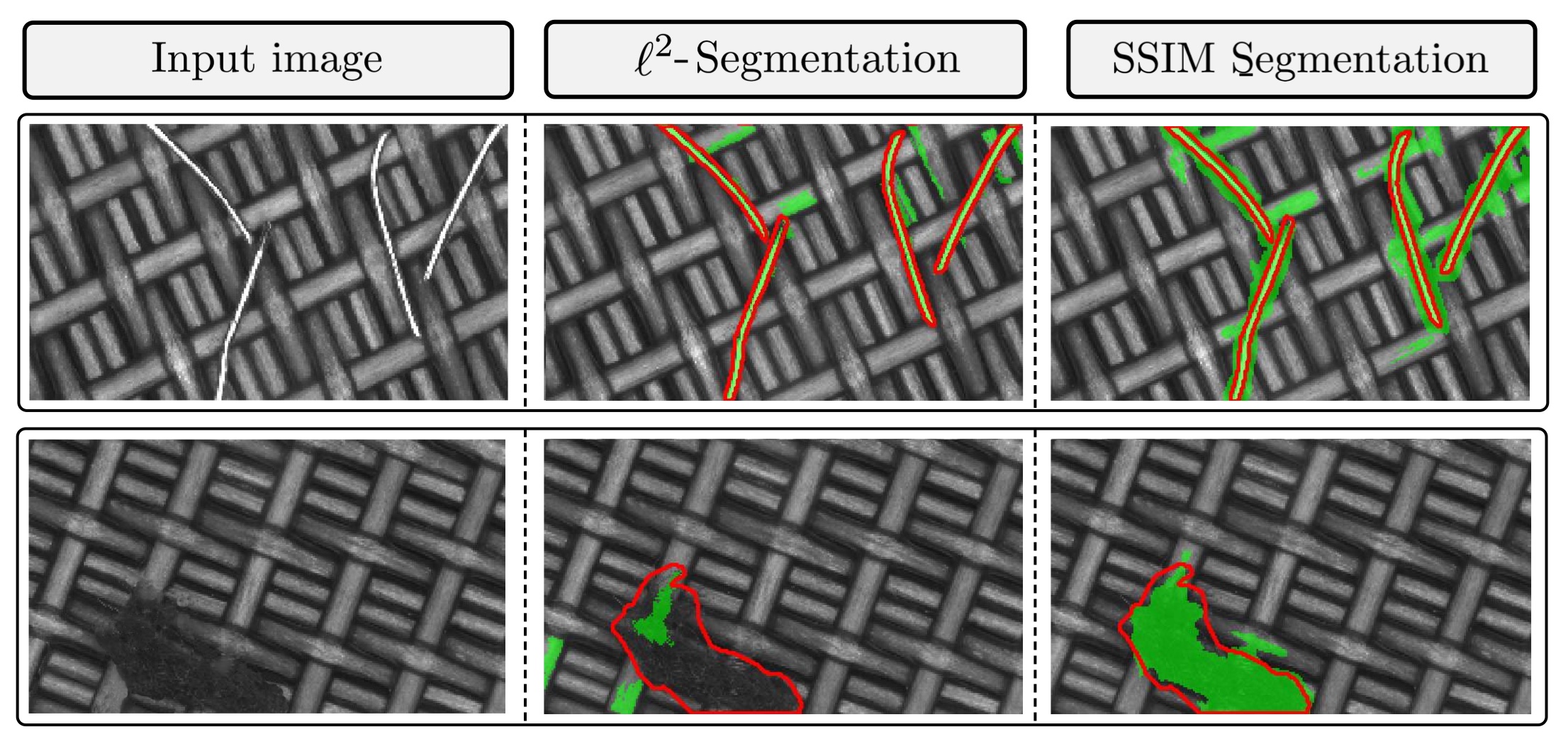}
  \caption{In the first row, the metal pins have a large difference in gray values in comparison to the defect-free background material. Therefore, they can be detected by both the $\ell^2$ and the SSIM error metric. The defect shown in the second row, however, differs from the texture more in terms of structure than in absolute gray values. As a consequence, a per-pixel distance metric fails to segment the defect while SSIM yields a good segmentation result.}
   \label{fig:metal_pin}
\end{figure}

\cref{fig:metal_pin} depicts qualitative advantages that employing a perceptual error metric has over per-pixel distances such as $\ell^2$. It displays two defective images from one of the texture datasets, where the top image contains a high-contrast defect of metal pins which contaminate the fabric. The bottom image shows a low-contrast structural defect where the fabric was cut open. While the $\ell^2$-norm has problems to detect the low-constrast defect, it easily segments the metal pins due to their large absolute distance in gray values with respect to the background. However, misalignments in edge regions still lead to large residuals in non-defective regions as well, which would make these thin defects hard to segment in practice. SSIM robustly segments both defect types due to its simultaneous focus on luminance, contrast, and structural information and insensitivity to edge alignment due to its patch-by-patch comparisons.

\section{\texorpdfstring{\uppercase{Conclusion}}{Conclusion}}
\label{sec:conclusion}

We demonstrate the advantage of perceptual loss functions over commonly used per-pixel residuals in autoencoding architectures when used for unsupervised defect segmentation tasks. Per-pixel losses fail to capture inter-dependencies between local image regions and therefore are of limited use when defects manifest themselves in structural alterations of the defect-free material where pixel intensity values stay roughly consistent. We further show that employing probabilistic per-pixel error metrics obtained by VAEs or sharpening reconstructions by feature matching regularization techniques do not improve the segmentation result since they do not address the problems that arise from treating pixels as mutually independent. 

SSIM, on the other hand, is less sensitive to small inaccuracies of edge locations due to its comparison of local patch regions and takes into account three different statistical measures: luminance, contrast, and structure. We demonstrate that switching from per-pixel loss functions to an error metric based on structural similarity yields significant improvements by evaluating on a challenging real-world dataset of nanofibrous materials and a contributed dataset of two woven fabric materials which we make publicly available. Employing SSIM often achieves an enhancement from almost unusable segmentations to results that are on par with other state of the art approaches for unsupervised defect segmentation which additionally rely on image priors such as pre-trained networks.

\begin{table}[t!]
\scriptsize
\centering
\begin{tabular}{cccccc}
\hline
\begin{tabular}[c]{@{}c@{}}Latent \\ dimension\end{tabular} & AUC & \begin{tabular}[c]{@{}c@{}}SSIM \\ window size\end{tabular} & AUC & Patch size & AUC \\ \hline
50                                                          & 0.848   & 3                                                           &  0.889   &            &    \\
100                                                         & 0.935   & 7                                                           &  0.965   & 32         & 0.949   \\
200                                                         & 0.961   & \textbf{11}                                                          &  \textbf{0.966}   & 64         & 0.959   \\
\textbf{500}                                                         & \textbf{0.966}   & 15                                                          &  0.960   & \textbf{128}        & \textbf{0.966}   \\
1000                                                        & 0.962   & 19                                                          &  0.952   &            &    \\ \hline
\end{tabular}
\caption{Area under the ROC curve (AUC) on NanoTWICE for varying hyperparameters in the SSIM autoencoder architecture. Different settings do not significantly alter defect segmentation performance.}
\label{table:hyperparams}
\end{table}

\bibliography{main}

\begin{thebibliography}{27}
\providecommand{\natexlab}[1]{#1}
\providecommand{\url}[1]{\texttt{#1}}
\expandafter\ifx\csname urlstyle\endcsname\relax
  \providecommand{\doi}[1]{doi: #1}\else
  \providecommand{\doi}{doi: \begingroup \urlstyle{rm}\Url}\fi

\bibitem[An and Cho(2015)]{vae_novelty_recon_probability}
Jinwon An and Sungzoon Cho.
\newblock {Variational Autoencoder based Anomaly Detection using Reconstruction
  Probability}.
\newblock \emph{SNU Data Mining Center, Tech. Rep.}, 2015.

\bibitem[Arjovsky and Bottou(2017)]{arjovsky2017towards_training_gans}
Martin Arjovsky and L{\'e}on Bottou.
\newblock {Towards Principled Methods for Training Generative Adversarial
  Networks}.
\newblock \emph{International Conference on Learning Representations}, 2017.

\bibitem[Baur et~al.(2018)Baur, Wiestler, Albarqouni, and
  Navab]{c_baur_vae_gan}
Christoph Baur, Benedikt Wiestler, Shadi Albarqouni, and Nassir Navab.
\newblock {Deep Autoencoding Models for Unsupervised Anomaly Segmentation in
  Brain MR Images}.
\newblock \emph{arXiv preprint arXiv:1804.04488}, 2018.

\bibitem[Boracchi et~al.(2014)Boracchi, Carrera, and
  Wohlberg]{sparse_representations_nanofibres}
Giacomo Boracchi, Diego Carrera, and Brendt Wohlberg.
\newblock {Novelty Detection in Images by Sparse Representations}.
\newblock In \emph{2014 IEEE Symposium on Intelligent Embedded Systems (IES)},
  pages 47--54. IEEE, 2014.

\bibitem[Carrera et~al.(2015)Carrera, Boracchi, Foi, and
  Wohlberg]{carrera_ano_conv_sparse}
Diego Carrera, Giacomo Boracchi, Alessandro Foi, and Brendt Wohlberg.
\newblock {Detecting anomalous structures by convolutional sparse models}.
\newblock In \emph{2015 International Joint Conference on Neural Networks
  (IJCNN)}, pages 1--8. IEEE, 2015.

\bibitem[Carrera et~al.(2016)Carrera, Boracchi, Foi, and
  Wohlberg]{carrera2016scale}
Diego Carrera, Giacomo Boracchi, Alessandro Foi, and Brendt Wohlberg.
\newblock {Scale-invariant anomaly detection with multiscale group-sparse
  models}.
\newblock In \emph{2016 IEEE International Conference on Image Processing
  (ICIP)}, pages 3892--3896. IEEE, 2016.

\bibitem[Carrera et~al.(2017)Carrera, Manganini, Boracchi, and
  Lanzarone]{handcrafted_feature_dictionary_nanofibres}
Diego Carrera, Fabio Manganini, Giacomo Boracchi, and Ettore Lanzarone.
\newblock {Defect Detection in {SEM} Images of Nanofibrous Materials}.
\newblock \emph{IEEE Transactions on Industrial Informatics}, 13\penalty0
  (2):\penalty0 551--561, 2017.

\bibitem[Donahue et~al.(2017)Donahue, Kr{\"a}henb{\"u}hl, and Darrell]{bigan}
Jeff Donahue, Philipp Kr{\"a}henb{\"u}hl, and Trevor Darrell.
\newblock {Adversarial Feature Learning}.
\newblock \emph{International Conference on Learning Representations}, 2017.

\bibitem[Dosovitskiy and Brox(2016)]{brox_learned_visual_similarity_metrics}
Alexey Dosovitskiy and Thomas Brox.
\newblock {Generating Images with Perceptual Similarity Metrics based on Deep
  Networks}.
\newblock In \emph{Advances in Neural Information Processing Systems}, pages
  658--666, 2016.

\bibitem[Goodfellow et~al.(2014)Goodfellow, Pouget-Abadie, Mirza, Xu,
  Warde-Farley, Ozair, Courville, and Bengio]{goodfellow2014generative}
Ian Goodfellow, Jean Pouget-Abadie, Mehdi Mirza, Bing Xu, David Warde-Farley,
  Sherjil Ozair, Aaron Courville, and Yoshua Bengio.
\newblock {Generative Adversarial Nets}.
\newblock In \emph{Advances in Neural Information Processing Systems}, pages
  2672--2680, 2014.

\bibitem[Goodfellow et~al.(2016)Goodfellow, Bengio, and
  Courville]{Goodfellow-et-al-2016}
Ian Goodfellow, Yoshua Bengio, and Aaron Courville.
\newblock \emph{{Deep Learning}}.
\newblock MIT Press, 2016.

\bibitem[Kingma and Ba(2015)]{kingma2014adam}
Diederik~P Kingma and Jimmy Ba.
\newblock {Adam: A Method for Stochastic Optimization}.
\newblock \emph{International Conference on Learning Representations}, 2015.

\bibitem[Kingma and Welling(2014)]{kingma2013vae}
Diederik~P Kingma and Max Welling.
\newblock {Auto-Encoding Variational Bayes}.
\newblock \emph{International Conference on Learning Representations}, 2014.

\bibitem[Krizhevsky et~al.(2012)Krizhevsky, Sutskever, and
  Hinton]{krizhevsky2012imagenet}
Alex Krizhevsky, Ilya Sutskever, and Geoffrey~E Hinton.
\newblock {ImageNet Classification With Deep Convolutional Neural Networks}.
\newblock In \emph{Advances in Neural Information Processing Systems}, pages
  1097--1105, 2012.

\bibitem[Napoletano et~al.(2018)Napoletano, Piccoli, and
  Schettini]{cnn_feature_dictionary_nanofibres}
Paolo Napoletano, Flavio Piccoli, and Raimondo Schettini.
\newblock {Anomaly Detection in Nanofibrous Materials by {CNN}-Based
  Self-Similarity}.
\newblock \emph{Sensors}, 18\penalty0 (1):\penalty0 209, 2018.

\bibitem[Perera and Patel(2018)]{perera_deep_features_one_class}
Pramuditha Perera and Vishal~M Patel.
\newblock {Learning Deep Features for One-Class Classification}.
\newblock \emph{arXiv preprint arXiv:1801.05365}, 2018.

\bibitem[Pimentel et~al.(2014)Pimentel, Clifton, Clifton, and
  Tarassenko]{pimentel2014review}
Marco~AF Pimentel, David~A Clifton, Lei Clifton, and Lionel Tarassenko.
\newblock {A review of novelty detection}.
\newblock \emph{Signal Processing}, 99:\penalty0 215--249, 2014.

\bibitem[Ridgeway et~al.(2015)Ridgeway, Snell, Roads, Zemel, and
  Mozer]{ae_vae_with_ms_ssim}
Karl Ridgeway, Jake Snell, Brett Roads, Richard~S Zemel, and Michael~C Mozer.
\newblock {Learning to generate images with perceptual similarity metrics}.
\newblock \emph{arXiv preprint arXiv:1511.06409}, 2015.

\bibitem[Russakovsky et~al.(2015)Russakovsky, Deng, Su, Krause, Satheesh, Ma,
  Huang, Karpathy, Khosla, Bernstein, et~al.]{russakovsky2015imagenet}
Olga Russakovsky, Jia Deng, Hao Su, Jonathan Krause, Sanjeev Satheesh, Sean Ma,
  Zhiheng Huang, Andrej Karpathy, Aditya Khosla, Michael Bernstein, et~al.
\newblock {ImageNet Large Scale Visual Recognition Challenge}.
\newblock \emph{International Journal of Computer Vision}, 115\penalty0
  (3):\penalty0 211--252, 2015.

\bibitem[Sabokrou et~al.(2018)Sabokrou, Khalooei, Fathy, and
  Adeli]{sabokrou2018adversarially_OCC}
Mohammad Sabokrou, Mohammad Khalooei, Mahmood Fathy, and Ehsan Adeli.
\newblock {Adversarially Learned One-Class Classifier for Novelty Detection}.
\newblock In \emph{Proceedings of the IEEE Conference on Computer Vision and
  Pattern Recognition}, pages 3379--3388, 2018.

\bibitem[Schlegl et~al.(2017)Schlegl, Seeb{\"o}ck, Waldstein, Schmidt-Erfurth,
  and Langs]{schlegl_anogan}
Thomas Schlegl, Philipp Seeb{\"o}ck, Sebastian~M Waldstein, Ursula
  Schmidt-Erfurth, and Georg Langs.
\newblock {Unsupervised Anomaly Detection with Generative Adversarial Networks
  to Guide Marker Discovery}.
\newblock In \emph{International Conference on Information Processing in
  Medical Imaging}, pages 146--157. Springer, 2017.

\bibitem[Soukup and Pinetz(2018)]{soukupreliably_disregard_decoder}
Daniel Soukup and Thomas Pinetz.
\newblock {Reliably Decoding Autoencoders’ Latent Spaces for One-Class
  Learning Image Inspection Scenarios}.
\newblock In \emph{OAGM Workshop 2018}. Verlag der Technischen Universit{\"a}t
  Graz, 2018.

\bibitem[Steger et~al.(2018)Steger, Ulrich, and
  Wiedemann]{machine_vision_algorithms_and_applications}
Carsten Steger, Markus Ulrich, and Christian Wiedemann.
\newblock \emph{Machine Vision Algorithms and Applications}.
\newblock Wiley-VCH, Weinheim, 2nd edition, 2018.

\bibitem[Vasilev et~al.(2018)Vasilev, Golkov, Lipp, Sgarlata, Tomassini, Jones,
  and Cremers]{q_space_novelty}
Aleksei Vasilev, Vladimir Golkov, Ilona Lipp, Eleonora Sgarlata, Valentina
  Tomassini, Derek~K Jones, and Daniel Cremers.
\newblock {q-Space Novelty Detection with Variational Autoencoders}.
\newblock \emph{arXiv preprint arXiv:1806.02997}, 2018.

\bibitem[Wang et~al.(2003)Wang, Simoncelli, and Bovik]{ms_ssim_index}
Zhou Wang, Eero~P Simoncelli, and Alan~C Bovik.
\newblock {Multiscale structural similarity for image quality assessment}.
\newblock In \emph{Record of the Thirty-Seventh Asilomar Conference on Signals,
  Systems and Computers}, volume~2, pages 1398--1402. IEEE, 2003.

\bibitem[Wang et~al.(2004)Wang, Bovik, Sheikh, and Simoncelli]{ssim_index}
Zhou Wang, Alan~C Bovik, Hamid~R Sheikh, and Eero~P Simoncelli.
\newblock {Image quality assessment: from error visibility to structural
  similarity}.
\newblock \emph{IEEE transactions on image processing}, 13\penalty0
  (4):\penalty0 600--612, 2004.

\bibitem[Zenati et~al.(2018)Zenati, Foo, Lecouat, Manek, and
  Chandrasekhar]{zenati2018BiGAN}
Houssam Zenati, Chuan~Sheng Foo, Bruno Lecouat, Gaurav Manek, and
  Vijay~Ramaseshan Chandrasekhar.
\newblock {Efficient {GAN}-Based Anomaly Detection}.
\newblock \emph{arXiv preprint arXiv:1802.06222}, 2018.

\end{thebibliography}
\bibliographystyle{plainnat}

\end{document}